\documentclass[letterpaper]{article}
\usepackage[preprint]{aaai2027}
\usepackage[hyphens]{url}
\usepackage{graphicx}
\usepackage{natbib}
\usepackage{caption}
\usepackage{subcaption} 
\usepackage{algorithm}
\usepackage{algorithmic}
\usepackage{booktabs}
\usepackage{multirow}
\usepackage{amsmath}
\usepackage{amssymb}
\usepackage{amsthm}
\usepackage{xspace}

\newcommand{\name}{RACER\xspace}

\newtheoremstyle{propnamed}{\topsep}{\topsep}{\itshape}{}{\bfseries}{.}{ }{\thmname{#1}\ \thmnumber{#2}\thmnote{. #3}}
\theoremstyle{propnamed}
\newtheorem{proposition}{Proposition}
\theoremstyle{plain}

\title{Disagree to Accelerate: Closing the Loop on Diffusion Feature Forecasts}

\author{
Yanchao Li\textsuperscript{\rm 1,\rm 2},
Jiaqing Xie\textsuperscript{\rm 2},
Ben Gao\textsuperscript{\rm 2},
Wanhao Liu\textsuperscript{\rm 2},\\
Yanbo Wang\textsuperscript{\rm 3},
T. Y. Tsui\textsuperscript{\rm 4},
Jinfei Liu\textsuperscript{\rm 5},
Yuqiang Li\textsuperscript{\rm 2}\corresponding,
Tianfan Fu\textsuperscript{\rm 1,\rm 2}\corresponding
}
\affiliations{
\textsuperscript{\rm 1}Nanjing University
\textsuperscript{\rm 2}Shanghai Artificial Intelligence Laboratory\\
\textsuperscript{\rm 3}North University of China
\textsuperscript{\rm 4}University of Pennsylvania
\textsuperscript{\rm 5}Zhejiang University
}

\newcommand{\projectpage}{\url{https://github.com/LiZaiyuan0619/RACER}}

\begin{document}
\maketitle

\begin{abstract}
Training-free feature forecasting accelerates diffusion sampling by predicting features at skipped denoising steps.
Recent work has mainly focused on designing stronger forecasters.
Yet forecast error varies sharply across steps, and open-loop caches trust the forecast in full at every skipped step.
This fixed trust is what breaks as acceleration turns aggressive.
The missing question is not only how to forecast better, but when and how much to trust a forecast.
We show that reliability can be observed from the cache itself.
Two forecasts agree where the feature trajectory is smooth, and they diverge where prediction turns hard.
Their disagreement is a cheap runtime signal, and it costs no extra denoiser evaluation.
Based on this signal, we introduce \name, a training-free closed-loop controller with two responses.
It continuously shrinks uncertain forecasts toward the last computed feature.
At the riskiest steps, \name refreshes the feature and repays the added evaluation by skipping a later scheduled one.
We derive a deterministic error bound for the shrinkage and empirically evaluate its validity and tightness across acceleration regimes.
At the same number of denoiser evaluations, \name improves the strongest open-loop baseline across SD3.5-Large, FLUX.1-dev, Wan2.1-14B, and HunyuanVideo on DrawBench, VBench, and COCO.
On SD3.5, we further show that \name samples faster at equal quality.
\name generalizes across forecasting designs as well.
For example, it recovers much of the quality lost on a Taylor base.
These results show that reliable diffusion acceleration also depends on how forecasts are used.
Code is available at \projectpage.
\end{abstract}

\section{Introduction}

Diffusion models dominate image and video generation \cite{NEURIPS2020_4c5bcfec,rombach2022high,peebles2023scalable}, but their sampling is slow. One sample runs a large denoiser tens to hundreds of times.
This number of function evaluations (NFE) drives the sampling time.
Training-free feature caching is a leading remedy \cite{Ma_2024_CVPR,selvaraju2024forafastforwardcachingdiffusion,Liu_2025_CVPR}.
It runs the network at only some steps and predicts the skipped features from a cache. Recent work has focused on improving the forecaster.
The rules have moved from local Taylor expansions to global fits such as Chebyshev polynomials \cite{Liu_2025_ICCV,Han_2026_CVPR}.
The remaining blind spot lies in how each forecast is used.
Whatever the rule, the prediction is used at full weight, with no check at run time.
Forecast error is far from uniform. It stays small on smooth stretches of the feature trajectory and spikes where it bends.
Even the strongest forecaster therefore becomes unreliable at high speedups. The challenge is then not the next forecasting rule. It is how far a given forecast should be trusted at each step.

\begin{figure*}[t]
  \centering
  \includegraphics[width=\textwidth]{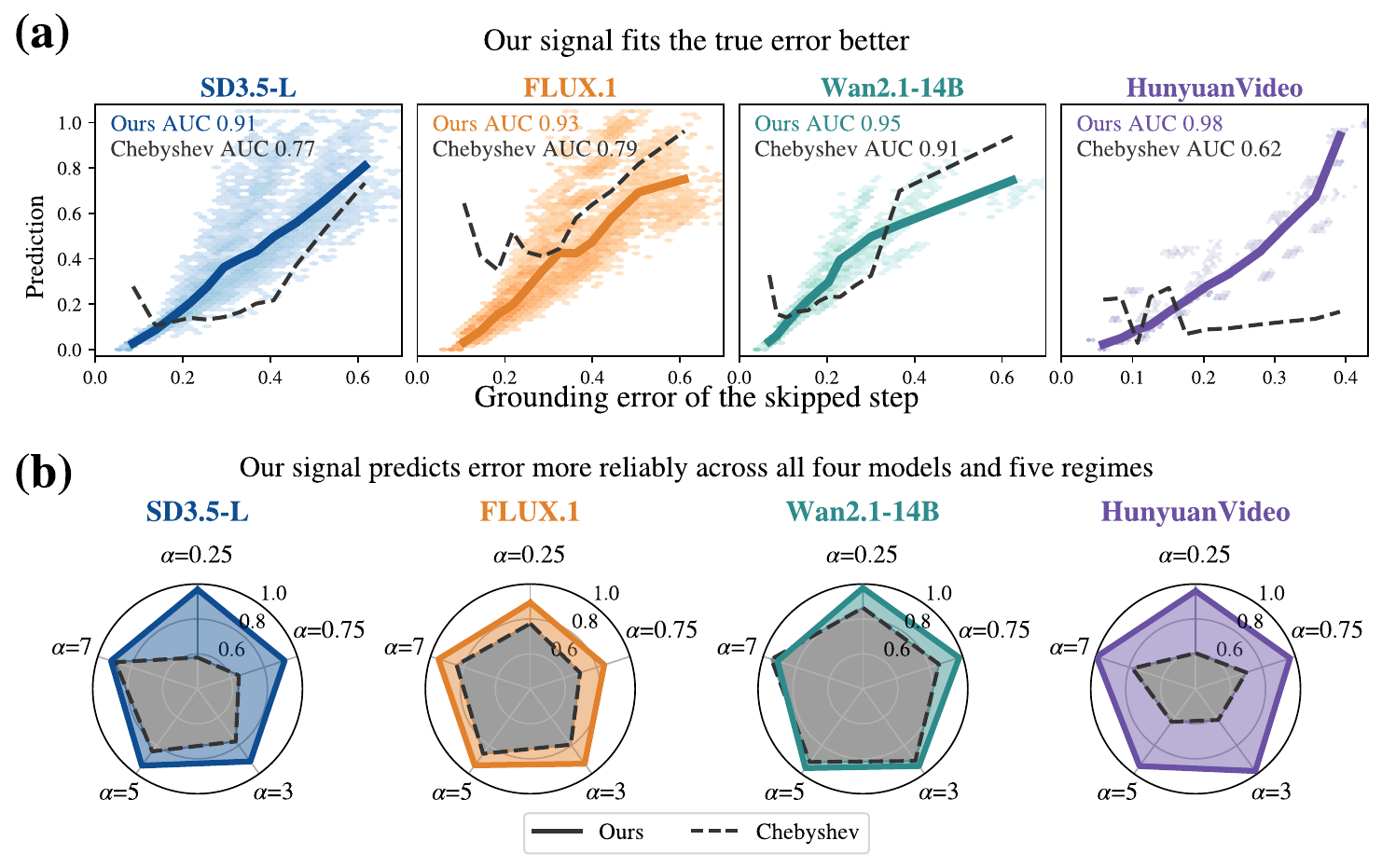}
  \caption{The disagreement $r$ is a cross-model reliability signal. (a) On four image and video models $r$ rises with the true forecast error, and it flags the high-error steps better than prior caches. The regime of this case is $\alpha$=3. (b) The signal separates the highest-error steps on every model and across all five acceleration regimes. By default, image models are evaluated on DrawBench, and video models on VBench.}
  \label{fig:hero}
\end{figure*}

The way a forecast fails leaves a trace in the cache.
Each cheap forecaster carries its own bias, so each drifts in its own way as prediction gets harder.
Two different forecasters therefore agree on the easy stretches and split apart on the hard ones.
The split is visible at run time, from the cache alone and with no extra denoiser evaluation.
We test this disagreement as a reliability signal on four image and video models.
It flags the highest-error steps at a mean AUROC of 0.94 (Figure~\ref{fig:hero}).
The input-side signal that prior caches rely on averages 0.77.

To fill the gap, we introduce \name, a Reliability-Aware Closed-loop controller with Exact Repay.
\name keeps the base forecaster unchanged. When the signal grows uncertain, \name shrinks the forecast continuously toward the last computed feature.
At a step too risky to trust at all, \name spends one denoiser evaluation to recompute the feature, then repays it by skipping a later scheduled step.
The per-prompt NFE therefore never moves from the base.
The shrinkage is grounded in a deterministic endpoint-error bound and an MSE-optimality analysis. We empirically evaluate the bound and its tightness on real traces.
A short offline trace fixes the controller scalars, with no training or denoiser modification.

At a matched NFE, \name improves the strongest open-loop baseline on all four models and all three benchmarks.
The advantage holds in every regime, and it is clearest at the aggressive end where the open-loop base breaks.
The margin also buys speed. On SD3.5, for instance, \name reaches a measured 5.4$\times$ over the 50-step sampler.
Even against the strongest base, it matches the base's best quality at $0.76\times$ the latency and $0.72\times$ the NFE.
Under a fixed refresh-only executor, the disagreement outperforms horizon and random signals by up to 2.42 dB.
The mechanism is also not tied to one forecaster pair. It rescues a Taylor base by up to 3.83 dB, and the observer itself can be swapped.
The controller form is shared across all four models, and image-side calibration transfers from SD3.5 to FLUX with only a 4.7\% change in the trust threshold.
We make four major contributions.

\begin{enumerate}
\item We recast feature caching as a question of the reliability of a forecast. This reliability is observable from the cache itself, as the disagreement between two cached forecasts.
\item We introduce \name, a training-free closed-loop controller that turns this signal into two responses, a continuous trust shrinkage and an exact refresh-and-repay.
\item The mechanism is decoupled from the forecaster. It generalizes to a new base forecaster and to a new observer, improving quality in both cases.
\item Across four image and video models and three benchmarks, \name improves the strongest open-loop baseline at matched NFE. On SD3.5, it also reaches equal quality at lower latency.
\end{enumerate}

\section{Related Work}

Diffusion sampling is slow because it runs the denoiser many times.
Two lines of work cut this cost.
One line takes fewer steps, with faster solvers, distillation, or straighter flow paths \cite{song2021denoising,lu2022dpmsolver,zhao2023unipc,salimans2022progressive,luo2023latentconsistencymodelssynthesizing,liu2022flowstraightfastlearning,lipman2023flow}.
The other makes each step cheaper, through quantization, pruning, or token merging, or spreads it across devices \cite{Li_2023_ICCV,fang2023structural,bolya2023tokenmergingfaststable,li2024distrifusiondistributedparallelinference}.
Feature caching belongs to the second line and needs no extra training \cite{Ma_2024_CVPR}.
Early caches reuse a stored feature, and recent work turns from reuse to forecasting \cite{zou2025accelerating}.
A forecaster reads the cache and extrapolates the skipped feature. Early rules used a local expansion \cite{Liu_2025_ICCV}.
A global basis fit came next, and its error grows more slowly with the skip horizon than a local expansion \cite{Han_2026_CVPR,trefethen2019approximation}.
Later work varies the basis, with Hermite, learned, or spectral forms \cite{zheng2026forecast}.
\name is orthogonal to these designs.
It treats the forecaster as a proposal and decides how much to trust it.

Most of these forecast caches apply the prediction open loop.
Once a step is skipped, the forecast is used in full, whatever its reliability.
A more recent group adds a runtime check, asking how far a cached prediction can be trusted~\cite{zheng2025compute}.
Forecast-then-verify caches recompute an actual feature and accept or reject the forecast.
Others read a shallow probe, an input-side change, or a feature change rate \cite{Liu_2025_CVPR,zhou2025enoughtrainingfreevideodiffusion,cui2026predictskiplinearmultistep}.
\name sits in this line and reads its signal on the output side, as the disagreement between two forecasts already in the cache.
It needs no extra denoiser evaluation and no input-side probe.
Prior signals mostly gate a binary recompute, while \name turns the signal into continuous trust.

\section{Method}

\subsection{Preliminaries}

A diffusion model draws a sample over a set of $N$ steps $T = \{t_1, \ldots, t_N\}$.
Each step evaluates the denoiser $\epsilon_\theta$ once, and a solver then advances the state,
\begin{equation}
x_i = \mathrm{Solve}\big(x_{i-1},\, \epsilon_\theta(x_{i-1}, t_i),\, t_i\big).
\label{eq:solve}
\end{equation}
One sample therefore runs the denoiser $N$ times, and this NFE dominates the sampling time.

Feature caching lowers the NFE by running the denoiser at only some steps and forecasting the rest \cite{Ma_2024_CVPR,selvaraju2024forafastforwardcachingdiffusion}.
The computed set $U \subseteq T$ holds the steps that run the denoiser and cache their features.
The skip set $V = T \setminus U$ holds the rest.
At a skipped step a forecaster $f$ predicts the missing feature from the cache, written $\hat{h}_t = f(C_t)$. Only $|U|$ steps run the denoiser, so this triple $(U, V, f)$ fixes the NFE.
Methods differ only in $f$, from feature reuse to local or global extrapolation.
We adopt a global Chebyshev forecaster as our base $f$ \cite{Han_2026_CVPR}. We call the nearest cached feature the anchor $a$.

Prior caches use this base open loop, trusting $\hat{h}_t$ in full at every skipped step.
But the forecast error is not uniform across steps.
A single fixed trust therefore breaks under aggressive skipping.

Our starting point is that the reliability of each forecast is readable from the cache alone, at no extra denoiser evaluation.
Alongside the base forecast $\hat{h}_t$, we compute a second forecast $\hat{g}_t = g(C_t)$ from the same cache.
We call this second forecaster the observer, and use a local Taylor rule for it.
Disagreement between predictors is a standard proxy for predictive uncertainty \cite{seung1992query,lakshminarayanan2017simple}.
The two forecasts carry different biases, so their disagreement is our reliability signal $r$,
\begin{equation}
r_t = \|\hat{h}_t - \hat{g}_t\| \,/\, \|\hat{h}_t\|.
\label{eq:signal}
\end{equation}
When the trajectory is smooth the two forecasts nearly agree, and $r$ stays small. Where the trajectory bends they diverge, and $r$ grows, marking a step where the forecast is unreliable.
Computing $r$ reuses the cache and adds only $O(F)$ arithmetic in the feature size $F$.
In the measured SD3.5 setting, this auxiliary computation adds about 2\% latency.
The effect is not tied to this pair, only to two forecasts with different biases.

We close the loop on $r$ with \name, a training-free controller.
It keeps the base $(U, V, f)$ unchanged and adds the observer $g$ and a control policy $\pi$.
The policy maps the runtime signal to two responses, $\pi: (r_t, k_t) \mapsto (\kappa_t, \gamma_t)$, where $\kappa_t$ is the trust placed on the forecast and $\gamma_t \in \{0, 1\}$ flags a refresh.
The method is thus the tuple $(U, V, f, g, \pi)$, orthogonal to the choice of $f$.
An open-loop cache is the special case in which neither output depends on $r_t$.

\begin{figure}[t]
  \centering
  \includegraphics[width=\columnwidth]{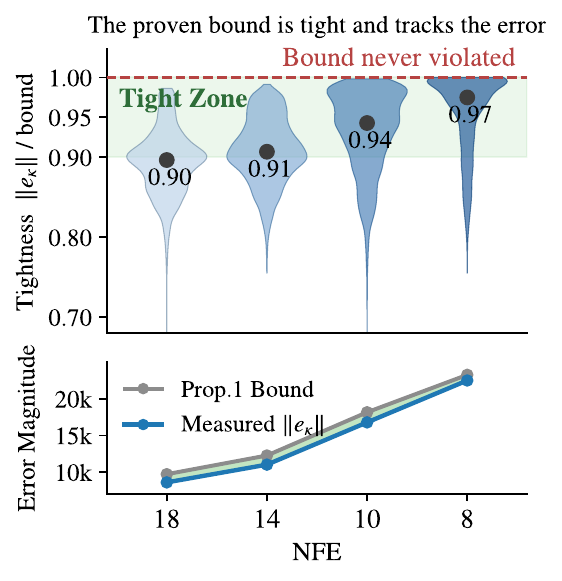}
  \caption{Empirical evaluation of Proposition~\ref{prop:bound}. The bound covers every evaluated step, remains tight with an error-to-bound ratio near 0.9--1.0, and tracks measured error across regimes.}
  \label{fig:theory}
\end{figure}

\subsection{Closed-Loop Control}

\begin{table*}[t]
\centering
\footnotesize
\setlength{\tabcolsep}{2pt}
\begin{tabular*}{\textwidth}{@{\extracolsep{\fill}} l l ccccc ccccc}
\toprule
 & & \multicolumn{5}{c}{SD3.5-Large} & \multicolumn{5}{c}{FLUX.1-dev} \\
\cmidrule(lr){3-7} \cmidrule(lr){8-12}
 & Method & PSNR$\uparrow$ & SSIM$\uparrow$ & LPIPS$\downarrow$ & ImageReward$\uparrow$ & CLIP$\uparrow$ & PSNR$\uparrow$ & SSIM$\uparrow$ & LPIPS$\downarrow$ & ImageReward$\uparrow$ & CLIP$\uparrow$ \\
\midrule
 & Reference & -- & -- & -- & 1.04 & 28.47 & -- & -- & -- & 1.01 & 27.38 \\
\midrule
\multirow{6}{*}{\shortstack[l]{$\alpha{=}0.25$\\ NFE 18}}
 & FORA & 10.84 & 0.485 & 0.591 & 0.85 & 28.33 & 16.17 & 0.673 & 0.387 & 0.99 & 27.32 \\
 & TaylorSeer & 11.49 & 0.547 & 0.505 & 0.81 & 27.96 & 19.87 & 0.781 & 0.227 & \underline{1.01} & \textbf{27.46} \\
 & TeaCache & 10.84 & 0.487 & 0.589 & 0.86 & 28.21 & 18.66 & 0.749 & 0.280 & 0.98 & \textbf{27.46} \\
 & ToCa & 12.79 & 0.586 & 0.449 & \underline{0.91} & \underline{28.46} & 10.67 & 0.450 & 0.665 & 0.99 & 27.34 \\
 & Chebyshev & \underline{18.35} & \underline{0.761} & \underline{0.232} & \textbf{1.01} & \textbf{28.48} & \underline{24.48} & \underline{0.849} & \textbf{0.139} & 1.00 & \underline{27.40} \\
 & Ours & \textbf{19.68} & \textbf{0.792} & \textbf{0.201} & \textbf{1.01} & \textbf{28.48} & \textbf{24.62} & \textbf{0.850} & \underline{0.142} & \textbf{1.03} & \textbf{27.46} \\
\midrule
\multirow{6}{*}{\shortstack[l]{$\alpha{=}0.75$\\ NFE${\approx}$14}}
 & FORA & 10.33 & 0.425 & 0.661 & 0.78 & \underline{28.47} & 15.24 & 0.638 & 0.442 & \underline{0.99} & 27.20 \\
 & TaylorSeer & 10.19 & 0.480 & 0.601 & 0.48 & 27.12 & 18.29 & 0.732 & 0.286 & 0.97 & \underline{27.40} \\
 & TeaCache & 11.85 & 0.555 & 0.513 & 0.91 & 28.26 & 18.71 & 0.751 & 0.277 & \textbf{1.03} & 27.34 \\
 & ToCa & 11.87 & 0.530 & 0.526 & 0.74 & 28.17 & 18.66 & 0.726 & 0.294 & 0.97 & 27.35 \\
 & Chebyshev & \underline{17.85} & \underline{0.737} & \underline{0.267} & \underline{0.97} & 28.38 & \underline{24.08} & \underline{0.837} & \textbf{0.146} & \underline{0.99} & 27.29 \\
 & Ours & \textbf{18.84} & \textbf{0.750} & \textbf{0.245} & \textbf{0.99} & \textbf{28.53} & \textbf{24.27} & \textbf{0.847} & \underline{0.164} & \textbf{1.03} & \textbf{27.41} \\
\midrule
\multirow{6}{*}{\shortstack[l]{$\alpha{=}3.0$\\ NFE${\approx}$10}}
 & FORA & 10.02 & 0.364 & 0.739 & 0.59 & 27.86 & 14.46 & 0.603 & 0.506 & 0.86 & 27.01 \\
 & TaylorSeer & 9.10 & 0.411 & 0.727 & $-$0.22 & 25.23 & 16.31 & 0.662 & 0.384 & 0.96 & 27.16 \\
 & TeaCache & 10.21 & 0.415 & 0.676 & 0.73 & \underline{28.04} & 16.75 & 0.681 & 0.380 & \textbf{1.03} & 27.32 \\
 & ToCa & 11.44 & 0.486 & 0.574 & 0.55 & 27.60 & 17.52 & 0.671 & 0.374 & 0.95 & 27.57 \\
 & Chebyshev & \underline{15.70} & \underline{0.613} & \underline{0.393} & \underline{0.79} & 27.73 & \underline{22.15} & \underline{0.772} & \textbf{0.234} & 0.99 & \underline{27.58} \\
 & Ours & \textbf{16.52} & \textbf{0.653} & \textbf{0.355} & \textbf{0.90} & \textbf{28.39} & \textbf{22.19} & \textbf{0.778} & \underline{0.252} & \underline{1.00} & \textbf{27.81} \\
\midrule
\multirow{6}{*}{\shortstack[l]{$\alpha{=}5.0$\\ NFE 9}}
 & FORA & 10.02 & 0.364 & 0.739 & 0.59 & 27.86 & 14.46 & 0.603 & 0.506 & 0.86 & 27.01 \\
 & TaylorSeer & 9.10 & 0.411 & 0.727 & $-$0.22 & 25.23 & 16.31 & 0.662 & 0.384 & 0.96 & 27.16 \\
 & TeaCache & 10.21 & 0.415 & 0.676 & \underline{0.73} & \underline{28.04} & 16.75 & 0.681 & 0.380 & \textbf{1.03} & 27.32 \\
 & ToCa & 11.44 & 0.486 & 0.574 & 0.55 & 27.60 & 17.52 & 0.671 & 0.374 & 0.95 & 27.57 \\
 & Chebyshev & \underline{15.06} & \underline{0.553} & \underline{0.465} & 0.47 & 26.83 & \underline{20.78} & \underline{0.717} & \underline{0.339} & 0.95 & \underline{27.76} \\
 & Ours & \textbf{15.73} & \textbf{0.618} & \textbf{0.397} & \textbf{0.83} & \textbf{28.14} & \textbf{21.44} & \textbf{0.742} & \textbf{0.299} & \underline{1.00} & \textbf{27.81} \\
\midrule
\multirow{6}{*}{\shortstack[l]{$\alpha{=}7.0$\\ NFE 8}}
 & FORA & 10.01 & 0.341 & 0.762 & \underline{0.48} & \underline{27.48} & 14.22 & 0.592 & 0.530 & 0.79 & 27.18 \\
 & TaylorSeer & 8.63 & 0.383 & 0.784 & $-$0.84 & 23.64 & 14.90 & 0.610 & 0.463 & 0.90 & 26.95 \\
 & TeaCache & 10.01 & 0.344 & 0.761 & 0.47 & 27.47 & 15.54 & 0.627 & 0.479 & 0.84 & 27.22 \\
 & ToCa & 11.14 & 0.450 & 0.632 & 0.05 & 26.79 & 10.93 & 0.435 & 0.696 & \underline{0.91} & 27.70 \\
 & Chebyshev & \underline{14.39} & \underline{0.494} & \underline{0.484} & 0.34 & 26.74 & \underline{20.06} & \underline{0.662} & \underline{0.405} & 0.87 & \underline{27.79} \\
 & Ours & \textbf{14.95} & \textbf{0.566} & \textbf{0.444} & \textbf{0.70} & \textbf{27.70} & \textbf{20.48} & \textbf{0.686} & \textbf{0.369} & \textbf{0.92} & \textbf{27.91} \\
\bottomrule
\end{tabular*}
\caption{Image main results on DrawBench, quality at matched NFE. Best in \textbf{bold}, second best \underline{underlined}. The four cache baselines land on one schedule for $\alpha{=}3.0$ and $5.0$, so their entries repeat there.}
\label{tab:main-image}
\end{table*}

The first component of $\pi$ shrinks trust continuously.
At a skipped step it outputs a feature between the anchor $a_t$ and the forecast $\hat{h}_t$, $\hat{y}_t = a_t + \kappa_t(\hat{h}_t - a_t)$.
The trust $\kappa_t$ falls as the horizon $k_t$ grows and as $r_t$ rises,
\begin{equation}
\kappa_t = \exp\!\big(-\lambda\,\max(k_t - 1, 0)\big)\cdot\sigma\!\big(\beta\,(\theta_\kappa - r_t)\big).
\label{eq:kappa}
\end{equation}
Here $\sigma$ is the sigmoid, and $k_t$ is the number of steps since the last computed step.
This response spends no denoiser evaluation, and only combines cached features.
Its two extremes recover existing methods, pure reuse at $\kappa_t = 0$ and the full forecast at $\kappa_t = 1$. \name lies between them, guided by $r$.

The second component of $\pi$ is an exact refresh-and-repay.
It sets $\gamma_t = 1$ where the disagreement exceeds a refresh threshold, $r_t > \theta_{\mathrm{ref}}$, marking a forecast too unreliable to trust even weakly.
At that step \name spends one real evaluation to recompute the true feature, a refresh.
It repays this evaluation by dropping a later scheduled step from a repayable subset $R \subseteq U$.
At the dropped step it uses the shrinkage output instead.
A debt counter tracks the unpaid refreshes, and \name refreshes only while a repayable step still lies ahead.
Every refresh therefore has a later step to pay it back.
The per-prompt count of real evaluations stays exactly at $|U|$.

A short offline trace fixes the trust scalars $\lambda$, $\beta$, $\theta_\kappa$ and the refresh threshold $\theta_{\mathrm{ref}}$.
It needs no training and does not modify the denoiser.
A fixed regime rule uses the step coordinate for $\alpha\leq0.75$ and the logSNR coordinate for $\alpha\geq3.0$.
The cross-model calibration protocol is specified in Experimental Setup.

\subsection{Theoretical Analysis}

\begin{table*}[t]
\centering
\footnotesize
\setlength{\tabcolsep}{4pt}
\begin{tabular*}{\textwidth}{@{\extracolsep{\fill}} l l cccc cccc}
\toprule
 & & \multicolumn{4}{c}{Wan2.1-14B} & \multicolumn{4}{c}{HunyuanVideo} \\
\cmidrule(lr){3-6} \cmidrule(lr){7-10}
 & Method & PSNR$\uparrow$ & SSIM$\uparrow$ & LPIPS$\downarrow$ & VBench-Quality$\uparrow$ & PSNR$\uparrow$ & SSIM$\uparrow$ & LPIPS$\downarrow$ & VBench-Quality$\uparrow$ \\
\midrule
 & Reference & -- & -- & -- & 83.13 & -- & -- & -- & 84.49 \\
\midrule
\multirow{6}{*}{$\alpha{=}0.75$}
 & FORA & 14.47 & 0.426 & 0.532 & 80.53 & 18.24 & 0.663 & 0.427 & 83.34 \\
 & TaylorSeer & 19.46 & 0.660 & 0.299 & \underline{82.66} & 24.73 & 0.805 & 0.243 & 84.03 \\
 & TeaCache & 19.13 & 0.628 & 0.347 & 82.59 & 23.88 & 0.783 & 0.270 & \underline{84.08} \\
 & ToCa & 16.88 & 0.537 & 0.410 & 82.18 & 21.26 & 0.732 & 0.357 & 83.79 \\
 & Chebyshev & \underline{23.01} & \underline{0.755} & \underline{0.158} & 82.41 & \underline{26.27} & \underline{0.814} & \textbf{0.125} & 83.14 \\
 & Ours & \textbf{23.80} & \textbf{0.773} & \textbf{0.151} & \textbf{82.74} & \textbf{26.81} & \textbf{0.847} & \underline{0.147} & \textbf{84.14} \\
\midrule
\multirow{6}{*}{$\alpha{=}3.0$}
 & FORA & 13.02 & 0.353 & 0.598 & 80.82 & 17.29 & 0.594 & 0.475 & 83.35 \\
 & TaylorSeer & 17.24 & 0.585 & 0.367 & 81.38 & 22.27 & 0.740 & 0.303 & 83.40 \\
 & TeaCache & 16.53 & 0.557 & 0.390 & \underline{81.85} & 21.42 & 0.725 & 0.334 & 81.71 \\
 & ToCa & 15.10 & 0.488 & 0.453 & 81.14 & 19.48 & 0.656 & 0.438 & \underline{83.62} \\
 & Chebyshev & \underline{21.85} & \underline{0.694} & \textbf{0.208} & 80.90 & \underline{24.42} & \underline{0.767} & \textbf{0.173} & 82.84 \\
 & Ours & \textbf{22.77} & \textbf{0.729} & \underline{0.215} & \textbf{82.14} & \textbf{25.01} & \textbf{0.799} & \underline{0.202} & \textbf{83.68} \\
\midrule
\multirow{2}{*}{$\alpha{=}5.0$}
 & Chebyshev & 21.60 & 0.666 & 0.272 & 79.90 & 23.78 & 0.747 & 0.231 & 80.77 \\
 & Ours & \textbf{22.34} & \textbf{0.707} & \textbf{0.248} & \textbf{81.66} & \textbf{23.81} & \textbf{0.752} & \textbf{0.228} & \textbf{81.31} \\
\midrule
\multirow{2}{*}{$\alpha{=}7.0$}
 & Chebyshev & 21.22 & 0.637 & 0.316 & 79.10 & 21.82 & 0.635 & 0.436 & 80.01 \\
 & Ours & \textbf{21.87} & \textbf{0.673} & \textbf{0.292} & \textbf{80.77} & \textbf{22.67} & \textbf{0.685} & \textbf{0.328} & \textbf{80.12} \\
\bottomrule
\end{tabular*}
\caption{Main results of video generation on VBench, quality at matched NFE. Best in \textbf{bold}, second best \underline{underlined}.}
\label{tab:main-video}
\end{table*}

We establish three complementary properties of the controller.
They provide a deterministic error bound for shrinkage, an MSE-optimal target for trust, and exact per-prompt budget conservation.

The shrinkage output interpolates the anchor and the forecast, so its error is controlled by the two endpoint errors.
\begin{proposition}[Error bound]
\label{prop:bound}
For any $\kappa \in [0, 1]$,
\[
\|\hat{y}_t(\kappa) - h_t\| \le \kappa\,B_f + (1 - \kappa)\,B_h,
\]
where $h_t$ is the true feature, $B_f$ bounds the base-forecast error, and $B_h$ bounds the drift from the anchor.
\end{proposition}
The result is distribution-free. On our traces the bound holds at every step, and the measured error averages 0.932 of it (Figure~\ref{fig:theory}).
It also preserves the ordering of the steps by error, at a Spearman correlation of 0.997.

The trust structure follows from an MSE-optimality argument.
\begin{proposition}[MSE-optimal shrinkage]
\label{prop:optkappa}
Treat the forecast and the anchor as two estimates of the true feature with finite error second moments.
The trust that minimizes the expected squared error of $\hat{y}_t(\kappa)$ over $\kappa \in [0,1]$ is
\[
\kappa^* = \Pi_{[0,1]}\!\left(
\frac{\sigma_h^2 - \rho_{fh}\,\sigma_f\sigma_h}{\sigma_f^2 + \sigma_h^2 - 2\,\rho_{fh}\,\sigma_f\sigma_h}
\right),
\]
where $\Pi_{[0,1]}$ denotes projection onto $[0,1]$, $\sigma_f^2$ and $\sigma_h^2$ are their error second moments, and $\rho_{fh}$ is their normalized error correlation.
\end{proposition}
This is the optimal combination of two correlated estimates \cite{wang2023forecast}, and it reduces to inverse-MSE weighting $\sigma_h^2 / (\sigma_f^2 + \sigma_h^2)$ when the two errors are uncorrelated.
As forecast risk grows relative to anchor risk, the optimal trust decreases.
Equation~(\ref{eq:kappa}) implements this structure with the horizon as a proxy for extrapolation risk and $r_t$ as a proxy for forecast uncertainty.
Its exponential--sigmoid map keeps trust bounded, while quantile calibration absorbs the signal scale.
On our traces the true error rises with $r$ within each horizon bucket, at rank correlations that are typically above 0.89.

This runtime surrogate requires $r$ to be a faithful proxy, which in turn depends on the observer being a valid check on the base.
Write the disagreement as $\Delta_t = \hat{h}_t - \hat{g}_t$. Two conditions make the check valid. 
First, a large disagreement forces a large forecast error $e_f$, since
\begin{equation}
e_f \ge \|\Delta_t\| - B_o,
\label{eq:obs1}
\end{equation}
whenever the observer error is bounded by $B_o$. Second, a large forecast error must surface in the disagreement, since
\begin{equation}
\|\Delta_t\| \ge e_f\,\sqrt{1 - \rho_{fo}^2},
\label{eq:obs2}
\end{equation}
when the forecast and observer errors have correlation $\rho_{fo}$, in the mean-square sense.
A good observer therefore has a small error and a low correlation with the forecast.
The Chebyshev and Taylor pair meets both.
On the worst steps their error correlation runs from 0.17 to 0.44, which keeps $\sqrt{1 - \rho_{fo}^2}$ between 0.90 and 0.98.

The refresh-and-repay policy provides a constructive exact-budget guarantee.
It spends a real evaluation on a flagged step and later drops a scheduled one to pay it back.
\begin{proposition}[Exact budget]
\label{prop:budget}
For every prompt, the number of denoiser evaluations equals $|U|$.
\end{proposition}
The debt grows only when a distinct repayable step still lies ahead, so it never exceeds the repayable steps that remain.
Every refresh is thus matched to one later skipped step, and the count holds by construction.
Across 20 configurations of 200 prompts each, it never deviated.
Together these results bound the error, characterize an MSE-optimal target for $\kappa$, and fix the budget.

\section{Experiments}

\subsection{Experimental Setup}

\noindent\textbf{Tasks. }
We evaluate on four diffusion models, two for images and two for video.
For image generation, we use SD3.5-Large \cite{esser2024scalingrectifiedflowtransformers} and FLUX.1-dev \cite{labs2025flux1kontextflowmatching} at 1024$\times$1024.
For video generation, we use Wan2.1-14B \cite{wan2025wanopenadvancedlargescale} and HunyuanVideo \cite{kong2025hunyuanvideosystematicframeworklarge}.
The reference for each model is its own 50-step sampler.
We benchmark on DrawBench \cite{saharia2022photorealistic} for image generation and VBench \cite{huang2023vbenchcomprehensivebenchmarksuite} for video generation.
We also test generalization to a second image dataset, COCO \cite{lin2015microsoftcococommonobjects}, in Table~\ref{tab:coco}.

\noindent\textbf{Baselines.} 
The strongest open-loop baseline is the Chebyshev forecaster.
We also compare against other baselines, FORA, TaylorSeer, TeaCache, and ToCa \cite{selvaraju2024forafastforwardcachingdiffusion,Liu_2025_ICCV,Liu_2025_CVPR,zou2025accelerating}.
Each baseline uses its published configuration.

\noindent\textbf{Evaluation metrics.}
We report reference fidelity using PSNR, SSIM \cite{wang2004image}, and LPIPS \cite{zhang2018unreasonableeffectivenessdeepfeatures} against the 50-step reference.
We separately report ImageReward \cite{xu2023imagerewardlearningevaluatinghuman} and CLIP \cite{radford2021learningtransferablevisualmodels} for images, and VBench-Quality for videos.
NFE measures denoiser computation, while measured latency and its speedup over the reference measure end-to-end speed.

\noindent\textbf{Implementation details.}
All methods are compared at a matched NFE. 
The per-prompt NFE of \name exactly equals that of its base, so every paired comparison uses the same number of full denoiser evaluations.
We sweep several acceleration regimes per model, from mild to aggressive.
Each regime fixes a target NFE, while the controller form is shared across models.
Image parameters are transferred from SD3.5 to FLUX, whereas video thresholds are calibrated per model and regime on a short trace.

\begin{table}[t]
\centering
\footnotesize
\setlength{\tabcolsep}{2pt}
\begin{tabular*}{\columnwidth}{@{} l l @{\extracolsep{\fill}} r r r r r @{}}
\toprule
 & & \multicolumn{5}{c}{$\alpha$} \\
\cmidrule(lr){3-7}
Model & Method & 0.25 & 0.75 & 3.0 & 5.0 & 7.0 \\
\midrule
\multirow{3}{*}{SD3.5-L} & Chebyshev & 17.53 & 17.08 & 15.02 & 14.38 & 13.82 \\
 & Ours & \textbf{18.70} & \textbf{17.96} & \textbf{15.79} & \textbf{15.11} & \textbf{14.33} \\
 & $\Delta$ & $+$1.17 & $+$0.88 & $+$0.77 & $+$0.73 & $+$0.51 \\
\midrule
\multirow{3}{*}{FLUX.1} & Chebyshev & 23.74 & 23.34 & 21.42 & 20.29 & 19.49 \\
 & Ours & \textbf{23.83} & \textbf{23.53} & \textbf{21.49} & \textbf{20.82} & \textbf{19.86} \\
 & $\Delta$ & $+$0.09 & $+$0.19 & $+$0.07 & $+$0.53 & $+$0.37 \\
\bottomrule
\end{tabular*}
\caption{Generalization to COCO, PSNR of the Chebyshev base and \name. $\Delta$ is the gain of Ours over the base. The gain transfers across regimes and models.}
\label{tab:coco}
\end{table}

\begin{figure}[t]
  \centering
  \includegraphics[width=\columnwidth]{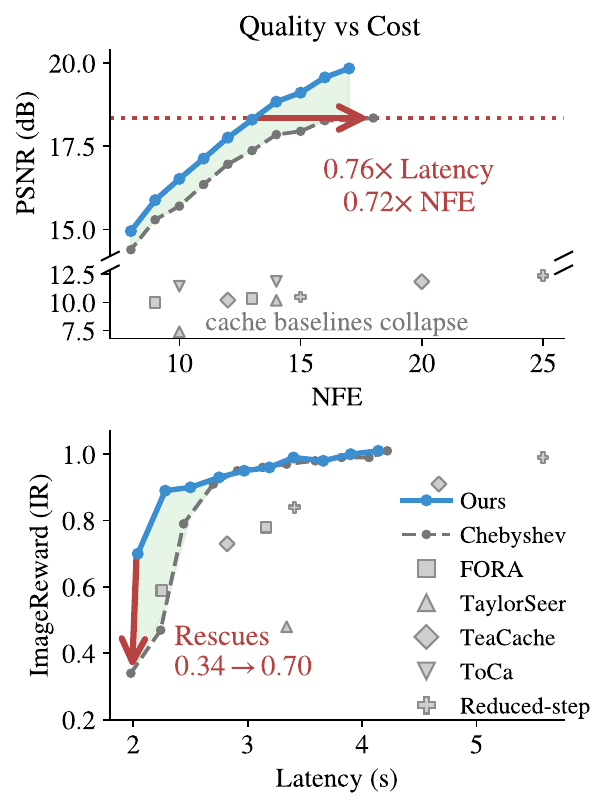}
  \caption{Quality versus compute on SD3.5. \name tops the Chebyshev base at every NFE and achieves the highest quality.}
  \label{fig:pareto}
\end{figure}

\subsection{Main Results}

\noindent\textbf{At matched NFE, \name improves PSNR and SSIM over the strongest open-loop baseline on all four models (Table~\ref{tab:main-image} and~\ref{tab:main-video}).}
The Chebyshev forecaster leads the other baselines on most metrics and regimes, and we treat it as the primary reference throughout.
The PSNR and SSIM gains hold in every regime, while the task metrics are matched or improved relative to the Chebyshev base.
LPIPS is the only metric with occasional losses.

\noindent\textbf{The gain also generalizes to a second image distribution (Table~\ref{tab:coco}).}
We evaluate on COCO to test how well the gain generalizes.
We compare against the Chebyshev base alone here.
\name raises the base in every regime.

\noindent\textbf{The quality margin converts to speed (Figure~\ref{fig:pareto}).}
For example, \name runs at a measured 2.6 to 5.4$\times$ speedup over the 50-step sampler across the sweep on SD3.5.
It tops the base at every point.

\begin{table}[t]
\centering
\footnotesize
\begin{tabular*}{\columnwidth}{@{} l l @{\extracolsep{\fill}} r r r r @{}}
\toprule
 & & \multicolumn{4}{c}{$\alpha$} \\
\cmidrule(lr){3-6}
Metric & Signal & 0.25 & 0.75 & 3.0 & 7.0 \\
\midrule
\multirow{3}{*}{PSNR$\uparrow$} & Ours & \textbf{19.38} & \textbf{18.13} & \textbf{15.50} & \textbf{14.35} \\
 & horizon & 17.56 & 15.71 & 14.72 & 13.42 \\
 & random & 17.90 & 16.30 & 14.04 & 13.16 \\
\midrule
\multirow{3}{*}{ImageReward$\uparrow$} & Ours & \textbf{1.01} & \textbf{0.97} & \textbf{0.81} & \textbf{0.28} \\
 & horizon & 0.99 & 0.78 & 0.63 & 0.03 \\
 & random & 0.98 & 0.85 & 0.35 & $-$0.29 \\
\bottomrule
\end{tabular*}
\caption{Signal ablation on SD3.5. The executor is the refresh alone, with the NFE and the trigger budget held fixed, so only the driving signal changes. Ours is the cache disagreement $r$.}
\label{tab:signal-abl}
\end{table}

\begin{table}[t]
\centering
\footnotesize
\begin{tabular*}{\columnwidth}{@{} l @{\extracolsep{\fill}} r r r r @{}}
\toprule
 & \multicolumn{4}{c}{$\alpha$} \\
\cmidrule(lr){2-5}
Method & 0.25 & 0.75 & 3.0 & 7.0 \\
\midrule
base & 18.35 & 17.85 & 15.70 & 14.39 \\
\midrule
refresh only & $+$1.03 & $+$0.28 & $-$0.20 & $-$0.04 \\
shrinkage only & $+$0.32 & $+$0.31 & $+$0.25 & $+$0.31 \\
Ours & \textbf{$+$1.54} & \textbf{$+$0.96} & \textbf{$+$0.49} & \textbf{$+$0.47} \\
\bottomrule
\end{tabular*}
\caption{Component ablation on SD3.5, under one configuration held fixed across regimes. The base row is the PSNR of the Chebyshev base, and the rows below it are the PSNR change over that base. Turning both responses on is best everywhere.}
\label{tab:comp-abl}
\end{table}

\subsection{Ablation Studies}

\noindent\textbf{The disagreement outperforms horizon and random signals under a fixed refresh-only executor (Table~\ref{tab:signal-abl}).}
We swap only the driving signal.
Replacing $r$ with the horizon or with a random rule loses quality in every regime.

\noindent\textbf{Both responses contribute, and they divide the work by regime (Table~\ref{tab:comp-abl}).}
On images, turning both on is best in every regime.
The refresh carries the mild regimes, and the shrinkage carries the aggressive ones, where the refresh alone no longer helps.

\begin{figure}[t]
  \centering
  \includegraphics[width=\columnwidth]{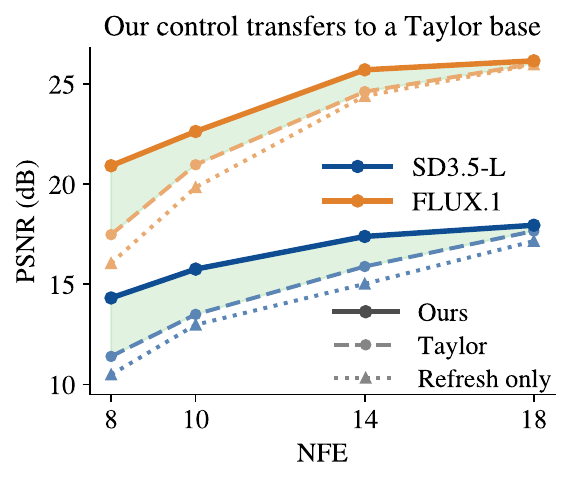}
  \caption{The \name controller placed on a Taylor base. The PSNR improvement over the Taylor base increases with acceleration.}
  \label{fig:transfer}
\end{figure}

\noindent\textbf{The controller also transfers to a Taylor base (Table~\ref{tab:taylor-base}, Figure~\ref{fig:transfer}).}
Here, a Taylor-internal observer drives the same control policy.
\name improves PSNR by up to 3.83 dB in the aggressive regimes.

\begin{table}[t]
\centering
\footnotesize
\setlength{\tabcolsep}{3pt}
\begin{tabular*}{\columnwidth}{@{} l l @{\extracolsep{\fill}} r r r r @{}}
\toprule
 & & \multicolumn{4}{c}{$\alpha$} \\
\cmidrule(lr){3-6}
Model & Method & 0.25 & 0.75 & 3.0 & 7.0 \\
\midrule
\multirow{3}{*}{SD3.5-L} & Taylor & 17.66 & 16.92 & 12.98 & 10.49 \\
 & refresh only & 17.16 & 15.02 & 13.50 & 11.40 \\
 & Ours & \textbf{17.95} & \textbf{17.39} & \textbf{15.76} & \textbf{14.32} \\
\midrule
\multirow{3}{*}{FLUX.1} & Taylor & 26.01 & 24.73 & 21.87 & 17.48 \\
 & refresh only & 25.97 & 24.40 & 20.84 & 16.65 \\
 & Ours & \textbf{26.16} & \textbf{25.71} & \textbf{22.63} & \textbf{20.92} \\
\bottomrule
\end{tabular*}
\caption{The \name controller on a Taylor base with PSNR.}
\label{tab:taylor-base}
\end{table}

\subsection{The Reliability Signal}

\noindent\textbf{The disagreement identifies high-error forecasts (Figure~\ref{fig:hero}).}
It flags the top-20\% forecast-error steps at a mean AUROC of 0.94, against 0.77 for the input-side signals that prior caches rely on (Figure~\ref{fig:hero}a).
The separation holds on every model and across all five regimes (Figure~\ref{fig:hero}b).

\begin{figure}[t]
  \centering
  \includegraphics[width=\columnwidth]{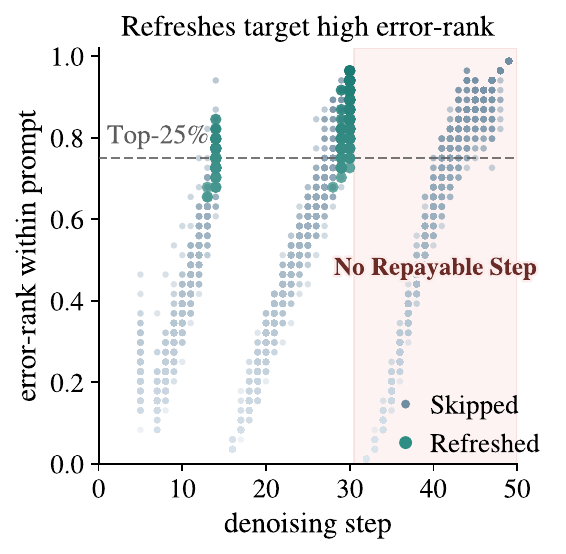}
  \caption{The signal places compute where it is needed. Refreshes are concentrated at high error ranks when later repayment is possible.}
  \label{fig:coloc}
\end{figure}

\noindent\textbf{Refreshes are concentrated at high-error steps (Figure~\ref{fig:coloc}).}
Wherever the budget allows, refreshes occur at high error ranks. The late steps carry no repayable step and stay forecast-only by design.

\begin{figure}[t]
  \centering
  \includegraphics[width=\columnwidth]{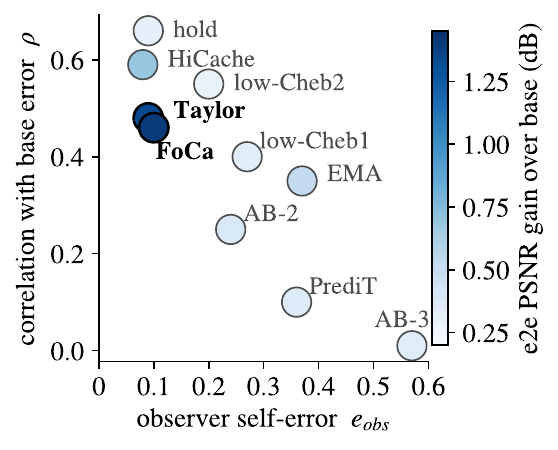}
  \caption{Observer choice on SD3.5 at $\alpha{=}0.25$. Taylor and FoCa yield the largest gains among ten observers. We use Taylor as the standard forecaster.}
  \label{fig:observer}
\end{figure}

\noindent\textbf{The Chebyshev and Taylor pair is a principled default (Figure~\ref{fig:observer}).}
The signal applies when the observer has low self-error and weak error correlation with the base.
Taylor and FoCa best balance these criteria, and Taylor-2 against Taylor-1 provides another effective pair.
The disagreement is a signal, not a better forecast.

\section{Conclusion}

This paper asked when and how much a diffusion feature forecast should be trusted. The answer sits in the cache itself.
Two cheap forecasts agree on the easy stretches and split where prediction turns hard, and this split flags the risky steps without an extra denoiser evaluation.
\name closes the loop on this signal with a continuous trust shrinkage and an exact refresh-and-repay.
At a matched NFE this improves the strongest open-loop base across four image and video generation models.
On SD3.5, \name also reaches equal quality faster.
The mechanism continues to work when either the base or the observer is replaced.
The signal scores how risky a step is, not which endpoint is better.
The gain is also smaller where the base already tracks the reference closely.
As forecasters grow stronger, the next lever for training-free acceleration may be reliable control under an exact budget.

\section*{Acknowledgments}
We thank Zhehong Ai for helpful discussions.

\bibliography{ref}

\end{document}